\documentclass[12pt,twoside]{article}
\usepackage{latexsym,amssymb}

\newtheorem{Def1}{Definition}
\newtheorem{Lemma}[Def1]{Lemma}
\newtheorem{Prop}[Def1]{Proposition}
\newtheorem{Theorem}[Def1]{Theorem}
\newtheorem{Cor}[Def1]{Corollary}

\newtheorem{Conj}[Def1]{Conjecture}

\newtheorem{Ex1}[Def1]{Example}
\newenvironment{Def} {\begin{Def1} \begin{upshape}} {\end{upshape} \end{Def1}}
\newenvironment{Example} {\begin{Ex1} \begin{upshape}} {\end{upshape} \end{Ex1}}

\newenvironment{Proof}[1][.]{\defaultskip\noindent \textbf{Proof}#1 }
  {\hspace*{0mm}\hfill $\Box$ \defaultskip}
\newenvironment{keywords}{\centerline{\bf\small
Keywords}\begin{quote}\small}{\par\end{quote}\vskip 1ex}
\def\defaultskip{\medskip}

\def\eins{1\hspace{-0.23em}{\rm I}}

\def\M{M}

\def\K{K\!}
\def\Kpre{K}
\def\Km{K\!m}

\def\beq{\begin{equation}}
\def\eeq{\end{equation}}
\def\beqn{\begin{displaymath}}
\def\eeqn{\end{displaymath}}
\def\bqa{\begin{eqnarray}}
\def\eqa{\end{eqnarray}}
\def\bqan{\begin{eqnarray*}}
\def\eqan{\end{eqnarray*}}

\def\NNN{\mathbb N}
\def\RRR{\mathbb R}
\def\QQQ{\mathbb Q}

\def\BBB{\mathbb B}

\def\Expect{{\mathbf E}}
\def\Prob{{\mathbf P}}

\def\eps{\varepsilon}

\def\equa{\stackrel+=}

\def\eqm{\stackrel\times=}
\def\leqm{\stackrel\times\leq}
\def\geqm{\stackrel\times\geq}

\def\leqn{_{1:n}}

\def\ltinf{_{<\infty}}
\def\_norm{_\mathrm{norm}}

\def\leqs#1{\stackrel {#1} \leq}

\def\for_all{\mbox{ for all }}
\def\such_that{\mbox{ such that }}

\def\und{\mbox{ and }}
\def\fuer{\mbox{ for }}

\def\lb{{\log_2}}                      
\def\l{\ell}

\def\ph{\varphi}
\def\th{\vartheta}

\begin{document}

\title{\normalsize\sc Technical Report \hfill IDSIA-13-04
\vskip 2mm\bf\Large\hrule height5pt \vskip 6mm
On the Convergence Speed of MDL Predictions for Bernoulli Sequences
\vskip 6mm \hrule height2pt \vskip 5mm}
\author{{\bf Jan Poland} and {\bf Marcus Hutter}\\[3mm]
\normalsize IDSIA, Galleria 2, CH-6928\ Manno-Lugano, Switzerland%
\thanks{This work was supported by SNF grant 2100-67712.02.}\\
\normalsize \{jan,marcus\}@idsia.ch, \ http://www.idsia.ch/$^{_{_\sim}}\!$\{jan,marcus\} }
\date{15 July 2004}
\maketitle

\begin{abstract}
We consider the Minimum Description Length principle for online
sequence prediction. If the underlying model class is discrete,
then the total expected square loss is a particularly interesting
performance measure: (a) this quantity is bounded, implying
convergence with probability one, and (b) it additionally
specifies a \emph{rate of convergence}. Generally, for MDL only
exponential loss bounds hold, as opposed to the linear bounds for
a Bayes mixture. We show that this is even the case if the model
class contains only Bernoulli distributions. We derive a new upper
bound on the prediction error for countable Bernoulli classes.
This implies a small bound (comparable to the one for Bayes
mixtures) for certain important model classes. The results apply
to many Machine Learning tasks including classification and
hypothesis testing. We provide arguments that our theorems
generalize to countable classes of i.i.d.\ models.
\end{abstract}

\newpage

\section{Introduction}\label{secIntro}

``Bayes mixture", ``Solomonoff induction", ``marginalization", all
these terms refer to a central induction principle: Obtain a
predictive distribution by integrating the product of prior and
evidence over the model class. In many cases however, the Bayes
mixture cannot be computed, and even a sophisticated approximation
is expensive. The MDL or MAP (maximum a posteriori) estimator is
both a common approximation for the Bayes mixture and interesting
for its own sake: Use the model with the largest product of prior
and evidence. (In practice, the MDL estimator is usually being
approximated too, in particular when only a local maximum is
determined.)

How good are the predictions by Bayes mixtures and MDL? This
question has attracted much attention. In the context of
prediction, arguably the most important quality measure is the
\emph{total} or cumulative \emph{expected loss} of a predictor.
A very common choice of loss function is the square loss.
Throughout this paper, we will study this quantity in an
\emph{online setup}.

Assume that the outcome space is finite, and the model class is
continuously parameterized. Then for Bayes mixture prediction, the
cumulative expected square loss is usually small but unbounded,
growing with $\log n$, where $n$ is the sample size
\cite{Clarke:90}. This corresponds to an \emph{instantaneous} loss
bound of
$\frac{1}{n}$. For the MDL predictor, the losses behave similarly
\cite{Rissanen:96,Barron:98} under appropriate conditions, in
particular with a specific prior. (Note that in order to do MDL
for continuous model classes, one needs to \emph{discretize} the
parameter space, e.g.\ \cite{Barron:91}.)

On the other hand, if the model class is discrete, then
Solomonoff's theorem \cite{Solomonoff:78,Hutter:01alpha} bounds
the cumulative expected square loss for the Bayes mixture
predictions finitely, namely by $\ln w_\mu^{-1}$, where $w_\mu$ is
the prior weight of the ``true" model $\mu$. The only necessary
assumption is that the true distribution $\mu$ is contained in the
model class. For the corresponding MDL predictions, we have shown
\cite{Poland:04mdl} that a bound of $w_\mu^{-1}$ holds. This is
exponentially larger than the Solomonoff bound, and it is sharp in
general. A finite bound on the total expected square loss is
particularly interesting:
\begin{enumerate}
\item It implies convergence of the predictive to the true probabilities
with probability one. In contrast, an instantaneous loss bound
which tends to zero implies only convergence in probability.
\item Additionally, it gives a \emph{convergence speed}, in the
sense that errors of a certain magnitude cannot occur too often.
\end{enumerate}
So for both, Bayes mixtures and MDL, convergence with probability
one holds, while the convergence rate is exponentially worse for
MDL compared to the Bayes mixture.

It is therefore natural to ask if there are model classes where
the cumulative loss of MDL is comparable to that of Bayes mixture
predictions. Here we will concentrate on the simplest possible
stochastic case, namely discrete Bernoulli classes (compare also
\cite{Vovk:97}). It might be surprising to discover that in
general the cumulative loss is still exponential. On the other
hand, we will give mild conditions on the prior guaranteeing a
small bound. We will provide arguments that these results
generalize to arbitrary i.i.d.\ classes. Moreover, we will see
that the instantaneous (as opposed to the cumulative) bounds are
always small ($\approx\frac{1}{n}$). This corresponds to the
well-known fact that the instantaneous square loss of the Maximum
Likelihood estimator decays as $\frac{1}{n}$ in the Bernoulli
case.

A particular motivation to consider discrete model classes arises
in Algorithmic Information Theory. From a computational point of
view, the largest relevant model class is the countable class of
all computable models (isomorphic to programs) on some fixed
universal Turing machine. We may study the corresponding Bernoulli
case and consider the countable set of computable reals in
$[0,1]$. We call this the \emph{universal setup}. The description
length $K(\th)$ of a parameter $\th\in[0,1]$ is then given by the
length of the shortest program that outputs $\th$, and a prior
weight may be defined by $2^{K(\th)}$.

Many Machine Learning tasks are or can be reduced to sequence
prediction tasks. An important example is classification. The task
of classifying a new instance $z_n$ after having seen
(instance,class) pairs
$(z_1,c_1),...,(z_{n-1},c_{n-1})$ can be phrased as to predict the
continuation of the sequence
$z_1c_1...z_{n-1}c_{n-1}z_n$. Typically the (instance,class)
pairs are i.i.d.

Our main tool for obtaining results is the Kullback-Leibler
divergence. Lemmata for this quantity are stated in Section
\ref{secKL}. Section \ref{secLB} shows that the exponential error
bound obtained in \cite{Poland:04mdl} is sharp in general. In
Section \ref{secUB}, we give an upper bound on the instantaneous
and the cumulative losses. The latter bound is small e.g.\ under
certain conditions on the distribution of the weights, this is the
subject of Section \ref{secUDW}. Section \ref{secUC} treats the
universal setup. Finally, in Section \ref{secDC} we discuss the
results and give conclusions.

\section{Kullback-Leibler Divergence}\label{secKL}

Let $\BBB=\{0,1\}$ and consider finite strings $x\in\BBB^*$ as
well as infinite sequences $x\ltinf\in\BBB^\infty$, with the first
$n$ bits denoted by $x_{1:n}$. If we know
that $x$ is generated by an i.i.d random variable, then
$P(x_i=1)=\th_0$ for all $1\leq i\leq\l(x)$ where $\l(x)$ is the
length of $x$. Then $x$ is called a Bernoulli sequence, and
$\th_0\in\Theta\subset[0,1]$ the {\it true parameter}. In the
following we will consider only countable $\Theta$, e.g.\ the set
of all computable numbers in $[0,1]$.

Associated with each $\th\in\Theta$, there is a {\it complexity}
or description length $\K w(\th)$ and a {\it weight} or
(semi)probability $w_\th=2^{-\K w(\th)}$. The complexity will
often but need not be a natural number. Typically, one assumes
that the weights sum up to at most one,
$\sum_{\th\in\Theta}w_\th\leq 1$. Then, by the Kraft inequality,
for all $\th\in\Theta$ there exists a prefix-code of length
$\K w(\th)$. Because of this correspondence, it is only a matter of
convenience if results are developed in terms of description
lengths or probabilities. We will choose the former way. We won't
even need the condition $\sum_\th w_\th\leq 1$ for most of the
following results. This only means that $\K w$ cannot be
interpreted as a prefix code length, but does not cause other
problems.

Given a set of distributions $\Theta\subset [0,1]$, complexities
$\big(\K w(\th)\big)_{\th\in\Theta}$, a true distribution
$\th_0\in\Theta$, and some observed string $x\in\BBB^*$, we
define an {\it MDL estimator}%
\footnote{\label{footnoteMDL}Precisely, we define a MAP (maximum a posteriori)
estimator. For two reasons, our definition might not be considered
as MDL in the strict sense. First, MDL is often associated with a
specific prior, while we admit arbitrary priors. Second and more
importantly, when coding some data
$x$, one can exploit the fact that once the parameter $\th^x$ is
specified, only data which leads to this $\th^x$ needs to be
considered. This allows for a description shorter than $\K
w(\th^x)$. Nevertheless, the
\emph{construction principle} is commonly termed MDL, compare
e.g.\ the ``ideal MDL" in \cite{Vitanyi:00}.}:
\beqn
  \th^x =\arg\max_{\th\in\Theta}\{ w_\th P(x|\th_0=\th)\}.
\eeqn
Here, $P(x|\th_0=\th)$ is the probability of observing $x$ if
$\th$ is the true parameter. Clearly,
$P(x|\th_0=\th)=\th^{\eins(x)}(1-\th)^{\l(x)-\eins(x)}$,
where $\eins(x)$ is the number of ones in $x$. Hence
$P(x|\th_0=\th)$ depends only on $\l(x)$ and
$\eins(x)$. We therefore see
\bqa\label{eq:argmax}\label{defthhat}
  \th^x\ =\ \th^{(\alpha,n)}
  & = & \arg\max_{\th\in\Theta}\{ w_\th
  \left(\th^{\alpha}(1-\th)^{1-\alpha}\right)^n\}\\
  \nonumber
  & = & \arg\min_{\th\in\Theta}\{n\!\cdot\!D(\alpha\|\th)+\K w(\th)\!\cdot\ln 2\},
\eqa
where $n=\l(x)$ and $\alpha:=\frac{\eins(x)}{\l(x)}$ is the {\it
observed fraction} of ones and
$$\textstyle D(\alpha\|\th)=\alpha\ln\frac\alpha\th+(1-\alpha)\ln\frac{1-\alpha}{1-\th}$$
is the Kullback-Leibler divergence. Let $\th,\tilde\th\in\Theta$
be two parameters, then it follows from (\ref{eq:argmax}) that in
the process of choosing the MDL estimator, $\th$ is being
preferred to $\tilde\th$ iff
\beq \label{eqBeat} n\big(D(\alpha\|\tilde\th) -
D(\alpha\|\th)\big)\geq \ln 2 \cdot \big(\K w(\th)-\K
w(\tilde\th)\big).
\eeq
In this case, we say that $\th$ \emph{beats} $\tilde\th$. It is
immediate that for increasing
$n$ the influence of the complexities on the selection of the
maximizing element decreases.
We are now interested in the {\it total expected square prediction
error} (or cumulative square loss) of the MDL estimator
$\sum_{n=1}^\infty \Expect(\th^{x\leqn}-\th_0)^2$.
In terms of \cite{Poland:04mdl}, this is the {\it static MDL
prediction} loss, which means that a predictor/estimator $\th^x$
is chosen according to the current observation $x$. The
\emph{dynamic} method on the other hand would consider both
possible continuations $x0$ and $x1$ and predict according to
$\th^{x0}$
\emph{and} $\th^{x1}$. In the following, we concentrate on static
predictions. They are also preferred in practice, since computing
only one model is more efficient.

Let $A_n=\big\{\frac{k}{n}:0\leq k\leq n\big\}$. Given the true
parameter $\th_0$ and some $n\in\NNN$, the {\it expectation} of a
function $f^{(n)}:\{0,\ldots,n\}\to\RRR$ is given by
\beq\label{defE}
  \Expect f^{(n)}= \sum_{\alpha\in A_n} p(\alpha|n) f(\alpha n)
  \mbox{, where }
  p(\alpha|n)={n \choose k}\Big(\th_0^\alpha (1-\th_0)^{1-\alpha}\Big)^n.
\eeq
(Note that the probability $p(\alpha|n)$ depends on $\th_0$, which
we do not make explicit in our notation.) Therefore,
\beq
  \sum_{n=1}^\infty \Expect(\th^{x\leqn}-\th_0)^2=
  \sum_{n=1}^\infty \sum_{\alpha\in A_n} p(\alpha|n)
  (\th^{(\alpha,n)}-\th_0)^2.
\eeq
Denote the relation
$f=O(g)$ by $f\leqm g$. Analogously define ``$\geqm$"
and ``$\eqm$". From \cite[Corollary 12]{Poland:04mdl}, we
immediately obtain the following result.

\begin{Theorem}
\label{th:Previous}
The cumulative loss bound
$\sum_n \Expect(\th^{x\leqn}-\th_0)^2\leqm 2^{\K w(\th_0)}$ holds.
\end{Theorem}

This is the ``slow" convergence result mentioned in the
introduction. In contrast, for a Bayes mixture, the total expected
error is bounded by $\K w(\th_0)$ rather than $2^{\K w(\th_0)}$
(see \cite{Solomonoff:78} or \cite[Th.1]{Hutter:01alpha}). An
upper bound on $\sum_n \Expect(\th^{x\leqn}-\th_0)^2$ is termed as
\emph{convergence in mean sum} and implies convergence
$\th^{x\leqn}\to\th_0$ with probability 1 (since otherwise the sum
would be infinite).

We now establish relations between the Kullback-Leibler divergence
and the quadratic distance. We call bounds of this type {\it
entropy inequalities}.

\begin{Lemma} \label{Lemma:EntropyIneq}
Let $\th,\tilde\th\in(0,1)$ and
$\th^*=\arg\min\{|\th-\frac{1}{2}|,|\tilde\th-\frac{1}{2}|\}$, i.e.\ $\th^*$
is the element from $\{\th,\tilde\th\}$ which is closer to
$\frac{1}{2}$. Then
\bqan
2\cdot(\th-\tilde\th)^2 \leqs {(i)} & D(\th\|\tilde\th) &
\leqs {(ii)} \mbox{$\frac{8}{3}$}(\th-\tilde\th)^2 \und\\
\frac{(\th-\tilde\th)^2}{2\th^*(1-\th^*)}\leqs {(iii)}
& D(\th\|\tilde\th) & \leqs {(iv)}
\frac{3(\th-\tilde\th)^2}{2\th^*(1-\th^*)}.
\eqan
Thereby, $(ii)$ requires
$\th,\tilde\th\in[\frac{1}{4},\frac{3}{4}]$, $(iii)$ requires
$\th,\tilde\th\leq\frac{1}{2}$, and $(iv)$ requires
$\th\leq\frac{1}{4}$ and $\tilde\th\in[\frac{\th}{3},3\th]$.
Statements $(iii)$ and $(iv)$ have symmetric counterparts for
$\th\geq\frac{1}{2}$.
\end{Lemma}

\begin{Proof}
The lower bound $(i)$, is standard, see e.g.\ \cite[p.
329]{Li:97}. In order to verify the upper bound $(ii)$, let
$f(\eta)=D(\th\|\eta)-\frac{8}{3}(\eta-\th)^2$. Then $(ii)$
follows from $f(\eta)\leq 0$ for
$\eta\in[\frac{1}{4},\frac{3}{4}]$. We have that $f(\th)=0$ and
$f'(\eta)=\frac{\eta-\th}{\eta(1-\eta)}-\frac{16}{3}(\eta-\th)$.
This difference is nonnegative if and only $\eta-\th\leq 0$ since
$\eta(1-\eta)\geq\frac{3}{16}$. This implies $f(\eta)\leq 0$.
Statements $(iii)$ and $(iv)$ giving bounds if $\th$ is close to
the boundary are proven similarly.
\end{Proof}

Lemma \ref{Lemma:EntropyIneq} $(ii)$ is sufficient to prove the
lower bound on the error in Proposition \ref{Prop:Counterex}. The
bounds $(iii)$ and $(iv)$ are only needed in the technical proof
of the upper bound in Theorem \ref{Theorem:UpperBound}, which will
be omitted. It requires also similar upper and lower bounds for
the absolute distance, and if the second argument of
$D(\cdot\|\cdot)$ tends to the boundary. The lemma remains valid
for the extreme cases $\th,\tilde\th\in\{0,1\}$ if the fraction
$\frac{0}{0}$ is properly defined. It is likely to generalize to arbitrary
alphabet, for $(i)$ this is shown in \cite{Hutter:01alpha}.

It is a well-known fact that the binomial distribution may be
approximated by a Gaussian. Our next goal is to establish upper
and lower bounds for the binomial distribution. Again we leave out
the extreme cases.

\begin{Lemma} \label{Lemma:BinomialBounds}
Let $\th_0\in(0,1)$ be the true parameter, $n\geq 2$ and $1\leq
k\leq n-1$, and $\alpha=\frac{k}{n}$. Then the following
assertions hold.
\bqan
(i) && p(\alpha|n)\leq
\frac{1}{\sqrt{2\pi\alpha(1-\alpha)n}}\exp\big(-nD(\alpha\|\th_0)\big),\\
(ii) && p(\alpha|n)\geq
\frac{1}{\sqrt{8\alpha(1-\alpha)n}}\exp\big(-nD(\alpha\|\th_0)\big).
\eqan
\end{Lemma}

The lemma is verified using Stirling's formula. The upper bound is
sharp for $n\to\infty$ and fixed $\alpha$. Lemma
\ref{Lemma:BinomialBounds} can be easily combined with Lemma
\ref{Lemma:EntropyIneq}, yielding Gaussian estimates for the
Binomial distribution. The following lemma is proved by simply
estimating the sums by appropriate integrals.

\begin{Lemma} \label{Lemma:IntegralEstimate}
Let $z\in\RRR^+$, then
\bqan
  (i) && \frac{\sqrt\pi}{2z^3}-\frac{1}{z\sqrt{2e}}
  \leq \sum_{n=1}^\infty \sqrt{n}\cdot\exp(-z^2n)
  \leq \frac{\sqrt\pi}{2z^3}+\frac{1}{z\sqrt{2e}} \und\\
  (ii) && \sum_{n=1}^\infty n^{-\frac{1}{2}}\exp(-z^2n) \leq
  \sqrt\pi/z.
\eqan
\end{Lemma}

\section{Lower Bound}\label{secLB}

We are now in the position to prove that even for Bernoulli
classes the upper bound from Theorem \ref{th:Previous} is sharp in
general.

\begin{Prop} \label{Prop:Counterex}
Let $\th_0=\frac{1}{2}$ be the true parameter generating sequences
of fair coin flips. Assume there are $2^N-1$ other parameters
$\th_1,\ldots,\th_{2^N-1}$ with $\th_k=\frac{1}{2}+2^{-k-1}$.
Let all complexities be equal, i.e.\ $\K w(\th_0)=\K
w(\th_1)=\ldots=\K w(\th_{2^N-1})=N$. Then
\beqn
  \sum_{n=1}^\infty \Expect (\th_0-\th^x)^2\geq \mbox{$\frac{1}{84}$}\big(2^N-5\big)
  \eqm 2^{\K w(\th_0)}.
\eeqn
\end{Prop}

\begin{Proof}
Recall that $\th^x=\th^{(\alpha,n)}$ the maximizing element for
some observed sequence $x$ only depends on the length $n$ and the
observed fraction of ones $\alpha$. In order to obtain an estimate
for the total prediction error $\sum_n\Expect (\th_0-\th^x)^2$,
partition the interval $[0,1]$ into $2^N$ disjoint intervals
$I_k$, such that $\bigcup_{k=0}^{2^N-1} I_k=[0,1]$.
Then consider the contributions for the observed fraction $\alpha$
falling in $I_k$ {\it separately}:
\beq
\label{eq:Ck}
C(k)= \sum_{n=1}^\infty \sum_{\alpha\in A_n\cap
I_k}p(\alpha|n)(\th^{(\alpha,n)}-\th_0)^2
\eeq
(compare (\ref{defE})). Clearly, $\sum_n\Expect
(\th_0-\th^x)^2=\sum_k C(k)$ holds. We define the partitioning
$(I_k)$ as
$I_0=[0,\frac{1}{2}+2^{-2^N})=[0,\th_{2^N-1})$,
$I_1=[\frac{3}{4},1]=[\th_1,1]$, and
\beqn I_k=[\th_k,\th_{k-1}) \for_all 2\leq k\leq 2^N-1.
\eeqn
Fix $k\in\{2,\ldots,2^N-1\}$ and assume $\alpha\in I_k$. Then
\beqn
\th^{(\alpha,n)}=\arg\min_\th\{nD(\alpha\|\th)+\K w(\th)\ln 2\}
=\arg\min_\th\{nD(\alpha\|\th)\}\in\{\th_k,\th_{k-1}\}
\eeqn
according to (\ref{eq:argmax}). So clearly
$(\th^{(\alpha,n)}-\th_0)^2\geq(\th_k-\th_0)^2=2^{-2k-2}$ holds. Since
$p(\alpha|n)$ decreases for increasing $|\alpha-\th_0|$, we have
$p(\alpha|n)\geq p(\th_{k-1}|n)$. The interval $I_k$ has length $2^{-k-1}$,
so there are at least $\lfloor n 2^{-k-1}\rfloor\geq n 2^{-k-1}-1$
observed fractions $\alpha$ falling in the interval. From
(\ref{eq:Ck}), the total contribution of $\alpha\in I_k$ can be
estimated by
\beqn
C(k)\geq\sum_{n=1}^\infty 2^{-2k-2}(n2^{-k-1}-1)p(\th_{k-1}|n).
\eeqn
Note that the terms in the sum even become negative for small $n$,
which does not cause any problems. We proceed with
\beqn
p(\th_{k-1}|n) \geq
\frac{1}{\sqrt{8\cdot 2^{-2}n}}\exp
\big[-nD\big(\mbox{$\frac{1}{2}$}+2^{-k}\|\mbox{$\frac{1}{2}$}\big)\big]
\geq
\frac{1}{\sqrt{2n}}\exp
\big[-n\mbox{$\frac{8}{3}$} 2^{-2k}\big]
\eeqn
according to Lemma \ref{Lemma:BinomialBounds} and Lemma
\ref{Lemma:EntropyIneq} $(ii)$. By Lemma
\ref{Lemma:IntegralEstimate} $(i)$ and $(ii)$, we have
\bqan
\sum_{n=1}^\infty \sqrt n\exp \big[-n\mbox{$\frac{8}{3}$} 2^{-2k}\big]
& \geq & \frac{\sqrt\pi}{2}\left(\frac{3}{8}\right)^{\frac{3}{2}}2^{3k}-
\frac{1}{\sqrt{2e}}\sqrt{\frac{3}{8}} 2^{k}\und\\
-\sum_{n=1}^\infty n^{-\frac{1}{2}}\exp \big[-n\mbox{$\frac{8}{3}$} 2^{-2k}\big]
& \geq & -\sqrt\pi \sqrt{\frac{3}{8}} 2^{k}.
\eqan
Considering only $k\geq 5$, we thus obtain
\bqan
C(k)& \geq &
\frac{1}{\sqrt{2}}\sqrt{\frac{3}{8}}2^{-2k-2}
\left[\frac{3\sqrt\pi}{16}2^{2k-1}-\frac{1}{\sqrt{2e}} 2^{-1}-\sqrt\pi 2^{k}
\right]
\\ & \geq &
\frac{\sqrt 3}{16}
\left[3\sqrt\pi 2^{-5}-\frac{1}{\sqrt{2e}} 2^{-2k-1}-\sqrt\pi 2^{-k}
\right]
 \geq
\frac{\sqrt{3\pi}}{8}2^{-5}-\frac{\sqrt 3}{16\sqrt{2e}}2^{-11}
>\frac{1}{84}.
\eqan
Ignoring the contributions for $k\leq 4$, this implies the
assertion.
\end{Proof}

This result shows that if the parameters and their weights are
chosen in an appropriate way, then the total expected error is of
order $w_0^{-1}$ instead of $\ln w_0^{-1}$. Interestingly, this
outcome seems to depend on the arrangement and the weights of the
{\it false} parameters rather than on the weight of the {\it true}
one. One can check with moderate effort that the proposition still
remains valid if e.g.\ $w_0$ is twice as large as the other
weights. Actually, the proof of Proposition \ref{Prop:Counterex}
shows even a slightly more general result, namely the same bound
holds when there are additional arbitrary parameters with larger
complexities. This will be used for Example \ref{Ex:Sensitive}.
Other and more general assertions can be proven similarly.

\section{Upper Bounds}\label{secUB}

Although the cumulative error may be large, as seen in the
previous section, the instantaneous error is always small.

\begin{Prop}\label{propSUB}
For $n\geq 3$, the expected instantaneous square loss is bounded:
\beqn
  \Expect (\th_0-\hat\th^{x_{1:n}})^2
    \leq \frac{(\ln 2)\K w(\th_0)}{2n}+\frac{\sqrt{2(\ln 2)\K w(\th_0)\ln n}}{n}+
    \frac{6\ln n}{n}.
\eeqn
\end{Prop}

\begin{Proof}
We give an elementary proof for the case
$\th_0\in(\frac{1}{4},\frac{3}{4})$ only. Like in the proof of
Proposition \ref{Prop:Counterex}, we consider the contributions of
different $\alpha$ separately. By Hoeffding's inequality,
$\Prob(|\alpha-\th_0|\geq\frac{c}{\sqrt n})\leq2e^{-2c^2}$
for any $c>0$. Letting $c=\sqrt{\ln n}$, the contributions by
these $\alpha$ are thus bounded by $\frac{2}{n^2}\leq\frac{\ln
n}{n}$.

On the other hand, for $|\alpha-\th_0|\leq\frac{c}{\sqrt n}$,
recall that $\th_0$ \emph{beats} any $\th$ iff (\ref{eqBeat})
holds. According to $\K w(\th)\leq 1$,
$|\alpha-\th_0|\leq\frac{c}{\sqrt n}$, and Lemma
\ref{Lemma:EntropyIneq} $(i)$ and $(ii)$, (\ref{eqBeat}) is
already implied by
$
|\alpha-\th|\geq\sqrt{\frac{\frac{1}{2}(\ln 2)\K
w(\th_0)+\frac{4}{3}c^2}{n}}.
$
Clearly, a contribution only occurs if $\th$ beats
$\th_0$, therefore if the opposite inequality holds.
Using $|\alpha-\th_0|\leq\frac{c}{\sqrt n}$ again and the triangle
inequality, we obtain that
\beqn(\th-\th_0)^2\leq\frac{5c^2+
\frac{1}{2}(\ln 2)\K w(\th_0)+\sqrt{2(\ln 2)\K w(\th_0)c^2}}{n}
\eeqn
in this case. Since we have chosen $c=\sqrt{\ln n}$, this implies
the assertion.
\end{Proof}

One can improve the bound in Proposition \ref{propSUB} to
$\Expect (\th_0-\hat\th^{x_{1:n}})^2\leqm \frac{\K w(\th_0)}{n}$
by a refined argument, compare \cite{Barron:91}. But the
high-level assertion is the same: Even if the cumulative upper
bound may tend to infinity, the instantaneous error converges
rapidly to 0. Moreover, the convergence speed depends on
$\K w(\th_0)$ as opposed to $2^{\K w(\th_0)}$. Thus $\hat\th$
tends to $\th_0$ rapidly in probability (recall that the assertion
is not strong enough to conclude almost sure convergence). The
proof does not exploit $\sum w_\th\leq 1$, but only
$w_\th\leq 1$, hence the assertion even holds for a maximum
likelihood estimator (i.e.\ $w_\th=1$ for all
$\th\in\Theta$). The theorem generalizes to i.i.d.\ classes. For
the example in Proposition \ref{Prop:Counterex}, the instantaneous
bound implies that the bulk of losses occurs very late. This does
\emph{not} hold for general (non-i.i.d.) model classes: The losses in
\cite[Example 9]{Poland:04mdl} grow linearly in the first $n$
steps.

We will now state our main positive result that upper bounds the
cumulative loss in terms of the negative logarithm of the true
weight and the {\it arrangement} of the false parameters. We will
only give the proof idea -- which is similar to that of
Proposition \ref{Prop:Counterex} -- and omit the lengthy and
tedious technical details.

Consider the cumulated sum square error
$\sum_n \Expect \big(\th^{(\alpha,n)}-\th_0\big)^2$. In order to upper bound this
quantity, we will partition the open unit interval $(0,1)$ into a
sequence of intervals $(I_k)_{k=1}^\infty$, each of measure
$2^{-k}$. (More precisely: Each $I_k$ is either an interval or a
union of two intervals.) Then we will estimate the contribution of
each interval to the cumulated square error,
\beqn
C(k)=\sum_{n=1}^\infty \sum_{\alpha\in A_n,\th^{(\alpha,n)}\in
I_k} p(\alpha|n) (\th^{(\alpha,n)}-\th_0)^2
\eeqn
(compare (\ref{defE}) and (\ref{eq:Ck})). Note that
$\th^{(\alpha,n)}\in I_k$ precisely reads $\th^{(\alpha,n)}\in
I_k\cap\Theta$, but for convenience we generally assume
$\th\in\Theta$ for all $\th$ being considered. This
partitioning is also used for $\alpha$, i.e. define the
contribution $C(k,j)$ of $\th\in I_k$ where $\alpha\in I_j$ as
\beqn
C(k,j)=\sum_{n=1}^\infty \sum_{\alpha\in A_n\cap
I_j,\th^{(\alpha,n)}\in I_k} p(\alpha|n)
(\th^{(\alpha,n)}-\th_0)^2.
\eeqn
We need to distinguish between
$\alpha$ that are located close to $\th_0$ and
$\alpha$ that are located far from $\th_0$. ``Close" will
be roughly equivalent to $j>k$, ``far" will be
approximately $j\leq k$.
So we get
$\sum_n \Expect \big(\th^{(\alpha,n)}-\th_0\big)^2
= \sum_k^\infty C(k) = \sum_k\sum_j C(k,j)$. In the proof,
\beqn
p(\alpha|n)\leqm
\big[n\alpha(1-\alpha)\big]^{-\frac{1}{2}}\exp\big[-nD(\alpha\|\th_0)\big]
\eeqn
is often applied, which holds by Lemma \ref{Lemma:BinomialBounds}
(recall that $f\leqm g$ stands for
$f=O(g)$). Terms like $D(\alpha\|\th_0)$, arising in
this context and others, can be further estimated using Lemma
\ref{Lemma:EntropyIneq}. We now give the constructions of
intervals $I_k$ and complementary intervals $J_k$.

\begin{figure}[t]\begin{center}
\begin{minipage}{0.7\textwidth}\setlength{\unitlength}{0.75\textwidth}
\begin{picture}(1.3,0.8)
\multiput(0.15,0.2)(0,0.2){4}{\vector(1,0){1.1}}
\multiput(0.2,0.18)(0,0.2){4}{\line(0,1){0.04}}
\multiput(1.2,0.18)(0,0.2){4}{\line(0,1){0.04}}
\put(0.2,0.77){\makebox(0,0)[t]{$0$}}
\put(0.2,0.57){\makebox(0,0)[t]{$0$}}
\put(1.2,0.77){\makebox(0,0)[t]{$1$}}
\put(1.2,0.57){\makebox(0,0)[t]{{$\frac{1}{2}$}}}
\multiput(0.7,0.18)(0,0.2){4}{\line(0,1){0.04}}
\put(0.3875,0.78){\line(0,1){0.04}}
\put(0.3875,0.77){\makebox(0,0)[t]{$\th_0=\frac{3}{16}$}}
\put(0.7,0.77){\makebox(0,0)[t]{$\frac{1}{2}$}}
\put(0.575,0.58){\line(0,1){0.04}}
\put(0.575,0.57){\makebox(0,0)[t]{$\th_0$}}
\put(0.7,0.57){\makebox(0,0)[t]{$\frac{1}{4}$}}
\put(0.45,0.825){\oval(0.49,0.02)[t]}
\put(0.95,0.825){\oval(0.49,0.02)[t]}
\put(0.45,0.84){\makebox(0,0)[b]{$J_1$}}
\put(0.95,0.84){\makebox(0,0)[b]{$I_1$}}

\put(0.325,0.625){\oval(0.24,0.02)[t]}
\put(1.075,0.625){\oval(0.24,0.02)[t]}
\put(0.7,0.625){\oval(0.49,0.02)[t]}
\put(0.7,0.64){\makebox(0,0)[b]{$J_2$}}
\put(0.325,0.64){\makebox(0,0)[b]{$I_2$}}
\put(1.075,0.64){\makebox(0,0)[b]{$I_2$}}
\put(0.45,0.58){\line(0,1){0.04}}
\put(0.95,0.58){\line(0,1){0.04}}
\put(0.45,0.57){\makebox(0,0)[t]{$\frac{1}{8}$}}
\put(0.95,0.57){\makebox(0,0)[t]{$\frac{3}{8}$}}

\put(0.7,0.37){\makebox(0,0)[t]{$\frac{1}{4}$}}
\put(0.2,0.37){\makebox(0,0)[t]{$\frac{1}{8}$}}
\put(1.2,0.37){\makebox(0,0)[t]{$\frac{3}{8}$}}
\put(0.45,0.38){\line(0,1){0.04}}
\put(0.45,0.37){\makebox(0,0)[t]{$\th_0$}}
\put(0.45,0.425){\oval(0.49,0.02)[t]}
\put(0.95,0.425){\oval(0.49,0.02)[t]}
\put(0.45,0.44){\makebox(0,0)[b]{$J_3$}}
\put(0.95,0.44){\makebox(0,0)[b]{$I_3$}}

\put(0.2,0.17){\makebox(0,0)[t]{$\frac{1}{8}$}}
\put(1.2,0.17){\makebox(0,0)[t]{$\frac{1}{4}$}}
\put(0.7,0.17){\makebox(0,0)[t]{$\th_0=\frac{3}{16}$}}
\put(0.325,0.225){\oval(0.24,0.02)[t]}
\put(1.075,0.225){\oval(0.24,0.02)[t]}
\put(0.7,0.225){\oval(0.49,0.02)[t]}
\put(0.7,0.24){\makebox(0,0)[b]{$J_4$}}
\put(0.325,0.24){\makebox(0,0)[b]{$I_4$}}
\put(1.075,0.24){\makebox(0,0)[b]{$I_4$}}
\put(0.45,0.18){\line(0,1){0.04}}
\put(0.95,0.18){\line(0,1){0.04}}
\put(0.45,0.17){\makebox(0,0)[t]{$\frac{5}{32}$}}
\put(0.95,0.17){\makebox(0,0)[t]{$\frac{7}{32}$}}

\put(0.15,0.8){\makebox(0,0)[r]{$k=1$:}}
\put(0.15,0.6){\makebox(0,0)[r]{$k=2$:}}
\put(0.15,0.4){\makebox(0,0)[r]{$k=3$:}}
\put(0.15,0.2){\makebox(0,0)[r]{$k=4$:}}
\end{picture}
\end{minipage}\begin{minipage}{0.25\textwidth}
\caption{Example of the first four intervals for $\th_0=\frac{3}{16}$.
We have an l-step, a c-step, an l-step and another c-step. All following
steps will be also c-steps.}\label{fig:Intervals}
\end{minipage}
\end{center}\end{figure}
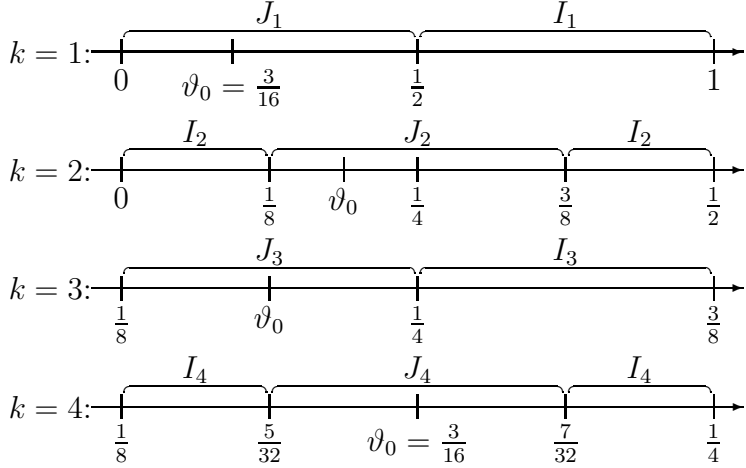

\begin{Def}
\label{Def:Intervals}
Let $\th_0\in\Theta$ be given.
Start with $J_0=[0,1)$. Let $J_{k-1}=[\th^l_k,\th^r_k)$
and define $d_k=\th^r_k-\th^l_k=2^{-k+1}$.
Then $I_k,J_k\subset J_{k-1}$ are constructed
from $J_{k-1}$ according to the following rules.
\bqa
\label{eq:Interval1}
\th_0\in[\th^l_k,\th^l_k+\mbox{$\frac{3}{8}$}d_k) & \Rightarrow &
J_k=[\th^l_k,\th^l_k+\mbox{$\frac{1}{2}$}d_k),\
I_k=[\th^l_k+\mbox{$\frac{1}{2}$}d_k,\th^r_k),\\
\label{eq:Interval2}
\th_0\in[\th^l_k+\mbox{$\frac{3}{8}$}d_k,\th^l_k+\mbox{$\frac{5}{8}$}d_k)
& \Rightarrow &
J_k=[\th^l_k+\mbox{$\frac{1}{4}$}d_k,\th^l_k+\mbox{$\frac{3}{4}$}d_k),
\\
\nonumber
&&
I_k=[\th^l_k,\th^l_k+\mbox{$\frac{1}{4}$}d_k)\cup
[\th^l_k+\mbox{$\frac{3}{4}$}d_k,\th^r_k),\\
\label{eq:Interval3}
\th_0\in[\th^l_k+\mbox{$\frac{5}{8}$}d_k,\th^r_k)
& \Rightarrow &
J_k=[\th^l_k+\mbox{$\frac{1}{2}$}d_k,\th^r_k),\
I_k=[\th^l_k,\th^l_k+\mbox{$\frac{1}{2}$}d_k).
\eqa
\end{Def}
We call the $k$th step of the interval construction an {\em
l-step} if (\ref{eq:Interval1}) applies, a {\em c-step} if
(\ref{eq:Interval2}) applies, and an {\em r-step} if
(\ref{eq:Interval3}) applies, respectively. Fig.\
\ref{fig:Intervals} shows an example for the interval
construction.

Clearly, this is not the only possible way to define an interval
construction. Maybe the reader wonders why we did not center the
intervals around $\th_0$. In fact, this construction would equally
work for the proof. However, its definition would not be easier,
since one still has to treat the case where
$\th_0$ is located close to the boundary. Moreover, our
construction has the nice property that the interval bounds are
finite binary fractions.
Given the interval construction, we can identify the $\th\in I_k$
with lowest complexity:
\bqan
\th^I_k & = & \arg\min\{\K w(\th):\th\in I_k\cap\Theta\},\\
\th^J_k & = & \arg\min\{\K w(\th):\th\in J_k\cap\Theta\},\und\\
\Delta(k) & = & \max\big\{\K w(\th^I_k)-\K w(\th^J_k),0\big\}.
\eqan
If there is no $\th\in I_k\cap\Theta$, we set $\Delta(k)=\K
w(\th^I_k)=\infty$.

\begin{Theorem}
\label{Theorem:UpperBound}
Let $\Theta\subset [0,1]$ be countable, $\th_0\in\Theta$, and
$w_\th=2^{-\K w(\th)}$, where $\K w(\th)$ is some complexity measure on $\Theta$.
Let $\Delta(k)$ be as introduced in the last paragraph, then
\beqn
\sum_{n=0}^\infty \Expect (\th_0-\th^x)^2\leqm \K w(\th_0)+
\sum_{k=1}^\infty 2^{-\Delta(k)}\sqrt{\Delta(k)}.
\eeqn
\end{Theorem}

The proof is omitted. But we briefly discuss the assertion of this
theorem. It states an error bound in terms of the arrangement of
the false parameters which directly depends on the interval
construction. As already indicated, a different interval
construction would do as well, provided that it exponentially
contracts to the true parameter. For a reasonable distribution of
parameters, we might expect that $\Delta(k)$ increases linearly
for $k$ large enough, and thus
$\sum 2^{-\Delta(k)}\sqrt{\Delta(k)}$ remains bounded. In the next
section, we identify cases where this holds.

\section{Uniformly Distributed Weights}\label{secUDW}

We are now able to state some positive results following from
Theorem \ref{Theorem:UpperBound}.

\begin{Theorem}
\label{Theorem:Th1}
Let $\Theta\subset[0,1]$ be a countable class of parameters and
$\th_0\in\Theta$ the true parameter. Assume that there are constants
$a\geq 1$ and $b\geq 0$ such that
\beq
\label{eq:Condition1}
\min\big\{\K w(\th):\th\in[\th_0-2^{-k},\th_0+2^{-k}]\cap\Theta,\th\neq\th_0\big\}\geq \frac{k-b}{a}
\eeq
holds for all $k>a\K w(\th_0)+b$. Then we have
\beqn
\sum_{n=0}^\infty \Expect (\th_0-\th^x)^2\leqm a \K w(\th_0)+b\leqm \K w(\th_0).
\eeqn
\end{Theorem}

\begin{Proof}
We have to show that
\beqn
\sum_{k=1}^\infty 2^{-\Delta(k)}\sqrt{\Delta(k)}\leqm a \K w(\th_0)+b,
\eeqn
then the assertion follows from Theorem \ref{Theorem:UpperBound}.
Let $k_1=\lceil a\K w(\th_0)+b+1 \rceil$ and $k'=k-k_1$. It is not
hard to see that $\max_{\th\in I_k}|\th-\th_0| \leq 2^{-k+1}$
holds. Together with (\ref{eq:Condition1}), this implies
\bqan
\sum_{k=1}^\infty 2^{-\Delta(k)}\sqrt{\Delta(k)}
& \leq & \sum_{k=1}^{k_1} 1 + \sum_{k=k_1+1}^\infty 2^{-\K
w(\th^I_k)+\K w(\th_0)}
\sqrt{\K w(\th^I_k)-\K w(\th_0)} \\
& \leq & k_1+2^{\K w(\th_0)}\sum_{k=k_1+1}^\infty 2^{-\frac{k-b}{a}}\sqrt{\frac{k-b}{a}}\\
& \leq & k_1+2^{\K w(\th_0)}\sum_{k'=1}^\infty 2^{-\frac{k'+k_1-b}{a}}\sqrt{\frac{k'+k_1-b}{a}}\\
& \leq & a\K w(\th_0)+b+2+\sum_{k'=1}^\infty
2^{-\frac{k'}{a}}\sqrt{\frac{k'}{a}+\K w(\th_0)}.
\eqan
Observe
$\sqrt{\frac{k'}{a}+\K w(\th_0)}\leq \sqrt{\frac{k'}{a}}+\sqrt{\K w(\th_0)}$,
$\sum_{k'}2^{-\frac{k'}{a}}\leqm a$, and by Lemma \ref{Lemma:IntegralEstimate}
$(i)$,
$\sum_{k'}2^{-\frac{k'}{a}}\sqrt{\frac{k'}{a}}\leqm a$.
Then the assertion follows.
\end{Proof}

Letting $j=\frac{k-b}{a}$, (\ref{eq:Condition1}) asserts that
parameters $\th$ with complexity $\K w(\th)=j$ must have a minimum
distance of $2^{-ja-b}$ from $\th_0$. That is, if parameters with
equal weights are (approximately) uniformly distributed in the
neighborhood of $\th_0$, in the sense that they are not too close
to each other, then fast convergence holds. The next two results
are special cases based on the set of all finite binary fractions,
\beqn
\QQQ_{\BBB^*}=\big\{\th=0.\beta_1\beta_2\ldots \beta_{n-1}1:n\in\NNN,\beta_i\in\BBB\big\}
\cup\big\{0,1\big\}.
\eeqn
If $\th=0.\beta_1\beta_2\ldots\beta_{n-1}1\in\QQQ_{\BBB^*}$, its
length is $\ell(\th)=n$. Moreover, there is a binary code
$\beta'_1\ldots\beta'_{n'}$ for $n$, having at most
$n'\leq\lfloor\lb (n+1)\rfloor$ bits. Then
$0\beta'_10\beta'_2\ldots 0\beta'_{n'}1\beta_1\ldots \beta_{n-1}$ is a
prefix-code for $\th$. For completeness, we can define the codes
for $\th=0,1$ to be $10$ and $11$, respectively. So we may define
a complexity measure on $\QQQ_{\BBB^*}$ by
\beq
\label{eq:ExampleCoding}
\K w(0)=2,\ \K w(1)=2, \und \K w(\th)=\ell(\th)+
2\big\lfloor\lb \big(\ell(\th)+1\big)\big\rfloor \fuer \th\neq
0,1.
\eeq
There are other similar simple prefix codes on $\QQQ_{\BBB^*}$
such that $\K w(\th)\geq \ell(\th)$.

\begin{Cor}
\label{Cor:Cor1}
Let $\Theta=\QQQ_{\BBB^*}$, $\th_0\in\Theta$ and $\K w(\th)\geq
\ell(\th)$, then
$\sum_n \Expect (\th_0-\th^x)^2\leqm \K w(\th_0)$
holds.
\end{Cor}

The proof is trivial, since Condition (\ref{eq:Condition1}) holds
with $a=1$ and $b=0$. This is a special case of a uniform
distribution of parameters with equal complexities. The next
corollary is more general, it proves fast convergence if the
uniform distribution is distorted by some function $\ph$.

\begin{Cor}
\label{Cor:Cor2}
Let $\ph:[0,1]\to[0,1]$ be an injective, $N$ times continuously
differentiable function. Let $\Theta=\ph(\QQQ_{\BBB^*})$,
$\K w\big(\ph(t)\big)\geq \ell(t)$ for all $t\in\QQQ_{\BBB^*}$, and
$\th_0=\ph(t_0)$ for a $t_0\in\QQQ_{\BBB^*}$. Assume that there is
$n\leq N$ and $\eps>0$ such that
\bqan
\left|\frac{d^n\ph}{dt^n}(t)\right|\ \geq \ c \ > & 0 & \for_all t\in[t_0-\eps,t_0+\eps]
\und\\
\frac{d^m\ph}{dt^m}(t_0)  = & 0 & \for_all 1\leq m<n.
\eqan
Then we have
\beqn
\sum \Expect (\th_0-\th^x)^2\leqm n\K w(\th_0)+2\lb(n!)-2\lb c+n\lb\eps \leqm n\K w(\th_0).
\eeqn
\end{Cor}

\begin{Proof}
Fix $j>\K w(\th_0)$, then
\beq
\label{eq:Kw11}
\K w\big(\ph(t)\big)\geq j \for_all
t\in[t_0-2^{-j},t_0+2^{-j}]\cap\QQQ_{\BBB^*}.
\eeq
Moreover, for all
$t\in[t_0-2^{-j},t_0+2^{-j}]$, Taylor's theorem asserts that
\beq
\label{eq:taylor}
\ph(t)=\ph(t_0)+\frac{\frac{d^n\ph}{dt^n}(\tilde t)}{n!}(t-t_0)^n
\eeq
for some $\tilde t$ in $(t_0,t)$ (or $(t,t_0)$ if $t<t_0$). We
request in addition $2^{-j}\leq\eps$, then
$|\frac{d^n\ph}{dt^n}|\geq c$ by assumption. Apply (\ref{eq:taylor}) to
$t=t_0+2^{-j}$ and $t=t_0-2^{-j}$ and define
$k=\lceil jn+\lb(n!)-\lb c\rceil$ in order to obtain
$|\ph(t_0+2^{-j})-\th_0|\geq 2^{-k}$ and
$|\ph(t_0-2^{-j})-\th_0|\geq 2^{-k}$. By injectivity of $\ph$,
we see that $\ph(t)\notin[\th_0-2^{-k},\th_0+2^{-k}]$ if
$t\notin[t_0-2^{-j},t_0+2^{-j}]$. Together with (\ref{eq:Kw11}), this implies
\beqn
\K w(\th)\geq j\geq\frac{k-\lb(n!)+\lb c-1}{n} \for_all
\th\in[\th_0-2^{-k},\th_0+2^{-k}]\cap\Theta.
\eeqn
This is condition (\ref{eq:Condition1}) with $a=n$ and
$b=\lb(n!)-\lb c+1$. Finally, the assumption $2^{-j}\leq\eps$
holds if $k\geq k_1=n\lb\eps+\lb(n!)-\lb c+1$. This gives an
additional contribution to the error of at most $k_1$.
\end{Proof}

Corollary \ref{Cor:Cor2} shows an implication of Theorem
\ref{Theorem:UpperBound} for {\it parameter identification}: A
class of models is given by a set of parameters $\QQQ_{\BBB^*}$
and a mapping $\ph:\QQQ_{\BBB^*}\to\Theta$. The task is to
identify the true parameter $t_0$ or its image $\th_0=\ph(t_0)$.
The injectivity of $\ph$ is not necessary for fast convergence,
but it facilitates the proof. The assumptions of Corollary
\ref{Cor:Cor2} are satisfied if $\ph$ is for example a polynomial.
In fact, it should be possible to prove fast convergence of MDL
for many common parameter identification problems. For sets of
parameters other than $\QQQ_{\BBB^*}$, e.g.\ the set of all
rational numbers
$\QQQ$, similar corollaries can easily be proven.

How large is the constant hidden in ``$\leqm$"? When examining
carefully the proof of Theorem \ref{Theorem:UpperBound}, the
resulting constant is quite large. This is mainly due to the
frequent ``wasting" of small constants. Supposably a smaller bound
holds as well, perhaps $16$. On the other hand, for the actual
{\it true} expectation (as opposed to its upper bound) and
complexities as in (\ref{eq:ExampleCoding}), numerical simulations
indicate that
$\sum_n \Expect (\th_0-\th^x)^2\leq \frac{1}{2}\K w(\th_0)$.

Finally, we state an implication which almost trivially follows
from Theorem \ref{Theorem:UpperBound}, since there
$\sum_k 2^{-\Delta(k)}\sqrt{\Delta(k)}\leq N$ is obvious. However,
it may be very useful for practical purposes, e.g. for hypothesis
testing.

\begin{Cor}
\label{Cor:Cor3}
Let $\Theta$ contain $N$ elements, $\K w(\cdot)$ be any complexity
function on $\Theta$, and $\th_0\in\Theta$. Then we have
\beqn
\sum_{n=1}^\infty \Expect (\th_0-\th^x)^2\leqm N+\K w(\th_0).
\eeqn
\end{Cor}

\section{The Universal Case} \label{secUC}

We briefly discuss the important universal setup, where $\K
w(\cdot)$ is (up to an additive constant) equal to the prefix
Kolmogorov complexity $\Kpre$ (that is the length of the shortest
self-delimiting program printing $\th$ on some universal Turing
machine). Since
$\sum_k 2^{-\Kpre(k)}\sqrt{\Kpre(k)}=\infty$ no matter how late the sum
starts (otherwise there would be a shorter code for large $k$), we
cannot apply Theorem \ref{Theorem:UpperBound}. This means in
particular that we do not even obtain our previous result, Theorem
\ref{th:Previous}. But probably the following strengthening of the
theorem holds under the same conditions, which then easily implies
Theorem \ref{th:Previous} up to a constant.

\begin{Conj}
\label{Conj:UpperBound}
$\sum_n \Expect (\th_0-\th^x)^2\leqm \Kpre(\th_0)+
\sum_k 2^{-\Delta(k)}$.
\end{Conj}

Then, take an incompressible finite binary fraction
$\th_0\in\QQQ_{\BBB^*}$, i.e.\
$\Kpre(\th_0)\equa \ell(\th_0)+\Kpre\big(\ell(\th_0)\big)$. For $k>\ell(\th_0)$, we can
reconstruct $\th_0$ and $k$ from $\th^I_k$ and $\ell(\th_0)$ by
just truncating $\th^I_k$ after $\ell(\th_0)$ bits. Thus
$\Kpre(\th^I_k)+\Kpre\big(\ell(\th_0)\big)\geqm
\Kpre(\th_0)+\Kpre\big(k|\th_0,\Kpre(\th_0)\big)$ holds. Using Conjecture
\ref{Conj:UpperBound}, we obtain
\beq\label{polylogbnd}
  \sum_n \Expect (\th_0-\th^x)^2
  \leqm \Kpre(\th_0)+2^{\Kpre(\ell(\th_0))}
  \leqm \ell(\th_0)\big(\lb \ell(\th_0)\big)^2,
\eeq
where the last inequality follows from the example coding given in
(\ref{eq:ExampleCoding}).
So, under Conjecture \ref{Conj:UpperBound}, we obtain a bound
which slightly exceeds the complexity $\Kpre(\th_0)$ if $\th_0$
has a certain structure. It is not obvious if the same holds for
all computable $\th_0$. In order to answer this question positive,
one could try to use something like \cite[Eq.(2.1)]{Gacs:83}. This
statement implies that as soon as $\Kpre(k)\geq K_1$ for all
$k\geq k_1$, we have $\sum_{k\geq k_1} 2^{-\Kpre(k)}\leqm
2^{-K_1}K_1(\lb K_1)^2$. It is possible to prove an analogous
result for $\th^I_k$ instead of $k$, however we have not found an
appropriate coding that does without knowing $\th_0$. Since the
resulting bound is exponential in the code length, we therefore
have not gained anything.

Another problem concerns the size of the multiplicative constant
that is hidden in the upper bound. Unlike in the case of uniformly
distributed weights, it is now of exponential size, i.e.\
$2^{O(1)}$. This is no artifact of the proof, as the following
example shows.

\begin{Example}
\label{Ex:Sensitive} Let $U$ be some universal Turing machine. We
construct a second universal Turing machine $U'$ from U as
follows: Let $N\geq 1$. If the input of $U'$ is $1^Np$, where
$1^N$ is the string consisting of $N$ ones and $p$ is some
program, then $U$ will be executed on $p$. If the input of $U'$ is
$0^N$, then $U'$ outputs $\frac{1}{2}$. Otherwise, if the input of
$U'$ is $x$ with $x\in\BBB^N\setminus\{0^N,1^N\}$, then $U'$
outputs $\frac{1}{2}+2^{-x-1}$. For $\th_0=\frac{1}{2}$, the
conditions of a slight generalization of Proposition
\ref{Prop:Counterex} are satisfied (where the complexity is
relative to $U'$), thus $\sum_n\Expect(\th^x-\th_0)^2\geqm 2^N$.
\end{Example}

Can this also happen if the underlying universal Turing machine is
not ``strange" in some sense, like $U'$, but ``natural"? Again
this is not obvious. One would have to define first a ``natural"
universal Turing machine which rules out cases like $U'$. If $N$
is not too large, then one can even argue that $U'$ {\it is}
natural in the sense that its compiler constant relative to $U$ is
small.

There is a relation to the class of all {\it deterministic}
(generally non-i.i.d.) measures. For this setup, MDL predicts the
next symbol just according to the {\it monotone complexity} $\Km$,
see \cite{Hutter:03unimdl}. According to \cite[Theorem
5]{Hutter:03unimdl}, $2^{-\Km}$ is very close to the universal
semimeasure $\M$ (this is due to \cite{Zvonkin:70}). Then the
total prediction error (which is defined slightly differently in
this case) can be shown to be bounded by
$2^{O(1)}\Km(x\ltinf)^3$ \cite{Hutter:04unimdlx}. The similarity to the
(unproven) bound (\ref{polylogbnd}) ``huge constant $\times$
polynomial" for the universal Bernoulli case is evident.

\section{Discussion and Conclusions}\label{secDC}

We have discovered the fact that the instantaneous and the
cumulative loss bounds can be \emph{incompatible}. On the one
hand, the cumulative loss for MDL predictions may be exponential,
i.e. $2^{\K w(\th_0)}$. Thus it implies almost sure convergence at
a slow rate, even for arbitrary discrete model classes
\cite{Poland:04mdl}. On the other hand, the instantaneous loss is
always of order
$\frac{1}{n}\K w(\th_0)$, implying fast convergence in probability
and a cumulative loss bound of $\K w(\th_0)\ln n$. Similar
logarithmic loss bounds can be found in the literature for
continuous model classes \cite{Rissanen:96}.

A different approach to assess convergence speed is presented in
\cite{Barron:91}. There in index of resolvability is introduced,
which can be interpreted as the difference of the expected MDL
code length and the expected code length under the true model. For
discrete model classes, they show that the index of resolvability
converges to zero as $\frac{1}{n}\K w(\th_0)$ \cite[Equation
(6.2)]{Barron:91}. Moreover, they give a convergence of the
predictive distributions in terms of the Hellinger distance
\cite[Theorem 4]{Barron:91}. This implies a cumulative (Hellinger)
loss bound of $\K w(\th_0)\ln n$ and therefore fast convergence in
probability.

If the prior weights are arranged nicely, we have proven a small
finite loss bound $\K w(\th_0)$ for MDL (Theorem
\ref{Theorem:UpperBound}). If parameters of equal complexity are
uniformly distributed or not too strongly distorted (Theorem
\ref{Theorem:Th1} and Corollaries), then the error is within a
small multiplicative constant of the complexity
$\K w(\th_0)$. This may be applied e.g.\ for the case of parameter
identification (Corollary \ref{Cor:Cor2}). A similar result holds
if $\Theta$ is finite and contains only few parameters (Corollary
\ref{Cor:Cor3}), which may be e.g.\ satisfied for hypothesis
testing. In these cases and many others, one can interpret the
conditions for fast convergence as the presence of prior
knowledge. One can show that if a predictor converges to the
correct model, then it performs also well under arbitrarily chosen
bounded loss-functions \cite[Theorem 4]{Hutter:02spupper}.
Moreover, we can then conclude good properties for other machine
learning tasks such as classification, as discussed in the
introduction. From an information theoretic viewpoint one may
interpret the conditions for a small bound in Theorem
\ref{Theorem:UpperBound} as ``good codes".

The main restriction of our positive result is the fact that we
have proved it only for the Bernoulli case. We therefore argue
that it generalizes to arbitrary i.i.d settings. Let
$\th_0\in[0,1]^N$, $\sum_i\th_0^{(i)}=1$ be a probability vector
that generates sequences of i.i.d.\ samples in
$\{1,\ldots,N\}^\infty$. Assume that $\th_0$ stays away from the
boundary (the other case is treated similarly). Then we can define
a sequence of nested sets in dimension $N-1$ in analogy to the
interval construction. The main points of the proof are now the
following two: First, for an observed parameter
$\alpha$ far from $\th_0$, the probability of $\alpha$ decays
exponentially, and second, for $\alpha$ close to $\th_0$, some
$\th$ far from $\th_0$ can contribute at most for short time.
These facts hold in the general i.i.d case like in the Bernoulli
case. However, the rigorous proof of it is yet more complicated
and technical than for the Bernoulli case. (Compare the proof of
the main result in \cite{Rissanen:96}.)

We conclude with an open question. In abstract terms, we have
proven a convergence result for the Bernoulli (or i.i.d) case by
mainly exploiting the {\it geometry} of the space of
distributions. This is in principle very easy, since for Bernoulli
this space is just the unit interval, for i.i.d it is the space of
probability vectors. It is not obvious how (or if at all) this
approach can be transferred to general (computable) measures.

\end{document}